\documentclass[journal]{IEEEtran}

\usepackage[utf8]{inputenc}
\usepackage{booktabs}
\usepackage{url}
\usepackage{color}
\usepackage{multirow}
\usepackage{cite}
\usepackage{amsmath,amssymb,amsfonts}
\usepackage{graphicx}
\usepackage{textcomp}
\usepackage{listings}
\usepackage{soul}
\sethlcolor{white}

\usepackage[noabbrev,capitalize]{cleveref}

\usepackage{color,array}

\usepackage{colortbl}
\usepackage[table]{xcolor} 
\newcommand*\rot{\rotatebox{90}}

\usepackage{adjustbox}  
\usepackage{afterpage}  

\usepackage{diagbox}

\usepackage{nidanfloat} 

\DeclareRobustCommand*{\IEEEauthorrefmark}[1]{%
  \raisebox{0pt}[0pt][0pt]{\textsuperscript{\footnotesize #1}}%
}

\usepackage{pifont}
\newcommand{\cmark}{\ding{51}}%
\newcommand{\xmark}{\ding{55}}%

\begin{document}

\markboth
{Preprint. Under review - Submitted November 6, 2020}
{Preprint. Under review - Submitted November 6, 2020}


\newcommand{\articleTitle}{FastPathology: An open-source platform for deep learning-based research and decision support in digital pathology}

\title{\articleTitle}



\author{\IEEEauthorblockN{Andr\'e Pedersen\IEEEauthorrefmark{a}\textsuperscript{,}\IEEEauthorrefmark{c}\textsuperscript{,}\IEEEauthorrefmark{$\ast$}, Marit Valla\IEEEauthorrefmark{a}\textsuperscript{,}\IEEEauthorrefmark{c}\textsuperscript{,}\IEEEauthorrefmark{d}, Anna M. Bofin\IEEEauthorrefmark{a}, Javier P\'erez de Frutos\IEEEauthorrefmark{e}, Ingerid Reinertsen\IEEEauthorrefmark{b}\textsuperscript{,}\IEEEauthorrefmark{e}, and Erik Smistad\IEEEauthorrefmark{b}\textsuperscript{,}\IEEEauthorrefmark{e}}
\thanks{This paper was submitted for review November 6, 2020. This work was supported by The Liaison Committee for Education, Research and Innovation in Central Norway (Samarbeidsorganet), and the Cancer Foundation, St. Olavs Hospital, Trondheim University Hospital (Kreftfondet).}
\thanks{\IEEEauthorblockA{\IEEEauthorrefmark{a}Department of Clinical and Molecular Medicine, The Norwegian University of Science and Technology, Trondheim, Norway}}
\thanks{\IEEEauthorblockA{\IEEEauthorrefmark{b}Department of Circulation and Imaging, The Norwegian University of Science and Technology, Trondheim, Norway}}
\thanks{\IEEEauthorblockA{\IEEEauthorrefmark{c}Clinic of Surgery, St. Olavs Hospital, Trondheim University Hospital, Trondheim, Norway}}
\thanks{\IEEEauthorblockA{\IEEEauthorrefmark{d}Department of Pathology, St. Olavs Hospital, Trondheim University Hospital, Trondheim, Norway}}
\thanks{\IEEEauthorblockA{\IEEEauthorrefmark{e}SINTEF Medical Technology, SINTEF, Trondheim, Norway}}
\thanks{\IEEEauthorblockA{\IEEEauthorrefmark{$\ast$}Corresponding author: andre.pedersen@ntnu.no}}
}

\maketitle

\begin{abstract} 
Deep convolutional neural networks (CNNs) are the current state-of-the-art for digital analysis of histopathological images. The large size of whole-slide microscopy images (WSIs) requires advanced memory handling to read, display and process these images. There are several open-source platforms for working with WSIs, but few support deployment of CNN models. These applications use third-party solutions for inference, making them less user-friendly and unsuitable for high-performance image analysis.
To make deployment of CNNs user-friendly and feasible on low-end machines, we have developed a new platform, \textit{FastPathology}, using the FAST framework and C\textit{++}. It minimizes memory usage for reading and processing WSIs, deployment of CNN models, and real-time interactive visualization of results.
Runtime experiments were conducted on four different use cases, using different architectures, inference engines, hardware configurations and operating systems.
Memory usage for reading, visualizing, zooming and panning a WSI were measured, using FastPathology and three existing platforms.
FastPathology performed similarly in terms of memory to the other C\textit{++}-based application, while using considerably less than the two Java-based platforms.
The choice of neural network model, inference engine, hardware and processors influenced runtime considerably.
Thus, FastPathology includes all steps needed for efficient visualization and processing of WSIs in a single application, including inference of CNNs with real-time display of the results.
Source code, binary releases and test data can be found online on GitHub at \url{https://github.com/SINTEFMedtek/FAST-Pathology/}.
\end{abstract}


\section{Introduction} \label{Introduction}

Whole Slide microscopy Images (WSIs) used in digital pathology are often large, and images captured at $\times400$ can have approximately $200k\times 100k$ color pixels resulting in an uncompressed size of $\sim 56$ GB \cite{Bandi2019}.
This exceeds the amount of RAM and GPU memory on most computer systems.
Thus, special data handling is required to store, read, process and display these images.

With the increasing integration of digital pathology into clinical practice worldwide, there is a need for tools that can assist clinicians in their daily practice.
Deep learning has shown great potential for automated and semi-automated analysis of medical images, including WSIs, with an accuracy surpassing traditional image analysis techniques \cite{LIU2019e271}.
Still, deploying Convolutional Neural Networks (CNNs) requires computer science expertise, making it difficult for clinicians and non-engineers to implement these methods into clinical practice.
Thus there is a need for an easy-to-use software that can load, visualize and process large WSIs using CNNs.

There are several open-source softwares available for visualizing and performing traditional image analysis on WSIs such as QuPath \cite{QuPath} and Orbit \cite{Orbit}.
Still, most of these do not support deployment of CNNs.
Most developers working with CNNs train their models in Python using frameworks like TensorFlow \cite{tensorflow2015} and Keras \cite{keras}. Thus, platforms intended for use in digital pathology should support deployment of these models.
A solution to this may be to deploy models in Python directly, using the same libraries, as done in Orbit. 
Inference is quite optimized in Python, because the actual Inference Engines (IEs), such as TensorFlow, are usually written in C and C\texttt{++}, and use parallel processing and GPUs.
The Python language itself is not optimized and is thus unfit for large scale, high performance software development. 
Most existing platforms use Java/Groovy as the main language. Despite boasting good multi-platform support and being a modern object-oriented language, the performance of Java compared to C and C\texttt{++} is debated \cite{Hundt2011,Gherardi2012}.
It is possible to deploy TensorFlow-based models in Java, with libraries like DeepLearning4J \cite{DL4J}, but its support for layers and network architectures is currently limited. 

We argue that due to the high memory and computational demands of processing and visualizing WSIs, modern C\texttt{++} together with GPU libraries such as OpenCL and OpenGL are better suited to create such a software.
We therefore propose to use and extend the existing high-performance C\texttt{++} framework FAST \cite{smistad2015fast} to develop an open-source platform for reading, visualizing and processing WSIs using deep CNNs.
FAST was introduced in 2015 as a framework for high performance medical image computing and visualization using multi-core CPUs and GPUs.
In 2019 \cite{8844665}, it was extended with CNN inference capabilities using multiple inference engines such as TensorFlow, OpenVINO \cite{OpenVINO} and TensorRT \cite{TensorRT}.
In this article, we describe a novel application \textit{FastPathology} based on FAST which consists of a Graphical User Interface (GUI) and open trained neural networks for analyzing digital pathology images.
We also outline the components that have been added to FAST to enable processing and visualization of WSIs.
Four different neural network inference cases, including patch-wise classification, low-resolution segmentation, high-resolution segmentation and object detection, are used to demonstrate the capabilities and computational performance of the platform. 
The application runs on both Windows and Ubuntu Linux Operating Systems (OSs) and is available online at \url{https://github.com/SINTEFMedtek/FAST-Pathology/}.

\subsection{Related Work}
\textbf{QuPath} \cite{QuPath} is a popular software for visualizing, analyzing and annotating WSIs. It is a Java-based application that supports reading WSIs using open-source readers such as Bio-Formats \cite{Bioformats2010} and OpenSlide \cite{OpenSlide}. QuPath can be applied directly using the GUI, but it also includes an integrated script editor for writing Groovy-based code for running more complex commands and algorithms.
Its annotation tool supports multiple different, dynamic brushes, and it can be used for various structures at different magnification levels. Using QuPath, it is possible to create new classifiers directly in the software, e.g. using Support Vector Machines (SVMs) and Random Forests (RFs). 
Quite recently, attempts to support deployment of trained CNNs have been made through StarDist \cite{stardist}, using TensorFlow to deploy a deep learning-based model for cell nucleus instance segmentation. Currently, the user cannot deploy their own trained CNNs in QuPath. However, it is possible to import external predictions from disk and save them as annotations.

The software \textbf{ASAP} \cite{ASAP} supports visualization, annotation and analysis of WSIs. Unlike QuPath, ASAP is based on C\texttt{++}. ASAP can also be used in Python directly through a wrapper, which is suitable as most machine learning researchers develop and train their models in Python. 

\textbf{Orbit} \cite{Orbit} is a recently released software. It includes processing and annotating tools similar to QuPath. However, it is possible to deploy and train CNNs directly in the software. Orbit is written in Java, but the deep learning-module is written in Python, and executed from Java. For computationally intensive tasks, such as training of CNNs, Orbit uses a Spark infrastructure, which makes it possible to relax the footprint on the local hardware.

Due to the large size of WSIs, utilizing algorithms on these has a high computational cost. \textbf{Cytomine} \cite{Cytomine} is a platform that solves this by running analyses through a web interface using a cloud-based service. It has similar options for visualization, annotation and analysis to QuPath. Its core solutions are open-source, however more advanced modules are not free-to-use. It also lacks options for CNN inference.

\section{Methods} \label{Design and implementation}
In the existing FAST framework, several components needed to be to be created to read, visualize and process WSIs. This section first describes how these components were designed to handle WSIs on a computer system with limited memory and computational resources.
Then, the FastPathology application itself is described, including how it was designed to enable users without programming experience to apply deep learning models on WSIs.

\subsection{Reading whole slide images} 
WSIs are usually stored in proprietary formats from various scanner vendors.
The open-source, C-based library OpenSlide \cite{OpenSlide} can read most of these proprietary formats.
Since these images are very large, they are usually stored as tiled image pyramids.
OpenSlide was added to FAST to enable reading of these files, thereby accessing the raw color pixel data.
OpenSlide uses the virtual memory mechanisms of the operating systems. Thus, by streaming data on demand from disk to RAM, it is possible to open and read large files without exhausting the RAM system memory.

\subsection{Creating arbitrary large images}
\label{sec:creating_large_images}
When performing image analysis tasks such as segmentation on high-resolution image planes of WSIs, it is necessary to create, write and read large images while performing segmentation using a sliding window approach. To facilitate this, a tiled image pyramid data object was added to FAST, enabling the creation of images of arbitrary sizes.
Given an image size of $M\times N$, FAST creates $L$ levels where each level has the size $\frac{M}{2^l} \times \frac{N}{2^l}$ with $l$ ranging from $0$ to $L-1$. Levels smaller than $4096\times 4096$ are not created.
Storing all levels in memory of a $\times400$ WSI, would require an extremely large amount of memory.
Thus the operating system of the native file-based memory mapping mechanisms are used, which on Linux is the 64 bit mmap function and on Windows the file mapping mechanism.
These file mapping mechanisms essentially create a large file on disk, and virtually map it to RAM, thus streaming data back and forth.
Reading and writing data in this manner is slower compared to using the RAM only. Furthermore, the speed is affected by disk speed, and it requires additional disk space.
To increase performance, levels that use less than an arbitrary threshold of 512 MB, are stored in RAM without memory mapping.

\subsection{Rendering a WSI with overlays}
High performance interactive image rendering with multiple overlays, colors and opacity usually requires a GPU implementation.
Since GPUs also have a very limited memory size, WSIs will not fit into the GPUs memory.
There is no native virtual memory system on GPUs, thus a virtual memory system for WSIs was implemented for GPUs in FAST, using OpenGL.
From the image pyramid, only the required tiles at the required resolution in the image pyramid are transferred to the GPU memory as textures.
To further reduce GPU memory usage, the tiles are stored using OpenGL's built-in texture compression algorithms.
The tiles and resolution required at a given time are automatically determined based on the current position and amount of zoom of the current view of the image.
Reading tiles from disk and streaming them to the GPU is time consuming. Therefore, the tiles are cached and put in a queue. 
The user can manually specify the maximum queue size in bytes.
Every time a tile is used, it is placed at the back of the queue. When the queue exceeds its limit, tiles are removed from the front of the queue and their textures deleted.
Before a given tile is ready to be rendered, the next-best resolution tiles already cached are displayed.
The lowest resolution of the image pyramid is always present in GPU memory. Thus, the WSI will always be displayed, even when higher resolution tiles are being loaded.

The user can easily pan and zoom to visualize all parts of a WSI with low latency and a bounded GPU memory usage.
In FAST, multiple images and objects can be displayed simultaneously with an arbitrary number of overlays. This enables high and low-resolution segmentations, patch-wise classifications, and bounding boxes to be displayed on top of a WSI with different colors and opacity levels. These can also be changed in real-time while processing.

\begin{figure*}[t!]
    \centering
    \includegraphics[width=1\textwidth]{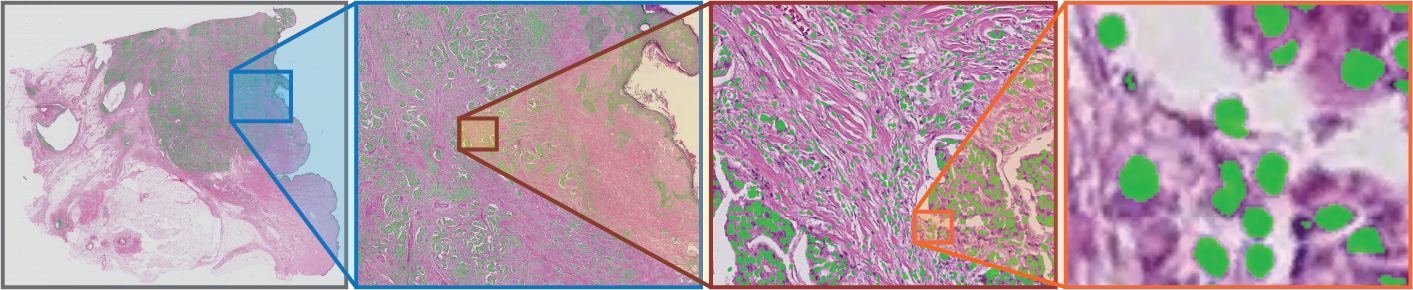}
    \caption{Illustration of how predictions, in this case segmented cell nuclei (green), can be visualized on top of a WSI in the viewer on different magnification levels.}
    \label{fig:zoom}
\end{figure*}

\cref{fig:zoom} shows an example of how the predictions can be visualized at different resolutions as overlays on top of the WSI. 
This illustrates the large size of these WSIs and why a tiled image pyramid data structure is required to visualize and process these images.



\subsection{Tissue segmentation}
Since WSIs are so large, applying a sliding window method across the image might be time consuming, especially when using CNNs. 
Thus, removing irrelevant regions such as glass would be an advantage.
In FAST, a simple tissue detector was implemented which segments the WSI by thresholding the RGB image color space. The image level with lowest resolution is segmented based on the Euclidean distance between a specific RGB triplet and the color white.
Morphological closing is then performed to bias sensitivity in tissue detection.
The default parameters were empirically determined and tuned on WSIs from a series of breast cancer tissue samples.
The tissue segmentation method was implemented in OpenCL to run in parallel on the GPU or on the multi-core CPU.

Otsu's method is commonly used to automatically set the threshold \cite{Oskal2019,Bandi2019_other,Guo2019}. However, it was observed that when the tissue section was large, covering almost the entire slide, the method produced thresholds that separated other tissue components, rather than background (glass). This phenomenon occurs because the method bases the threshold solely on the intensity histogram.

\begin{figure*}[t!]
    \centering
    \includegraphics[width=1\textwidth]{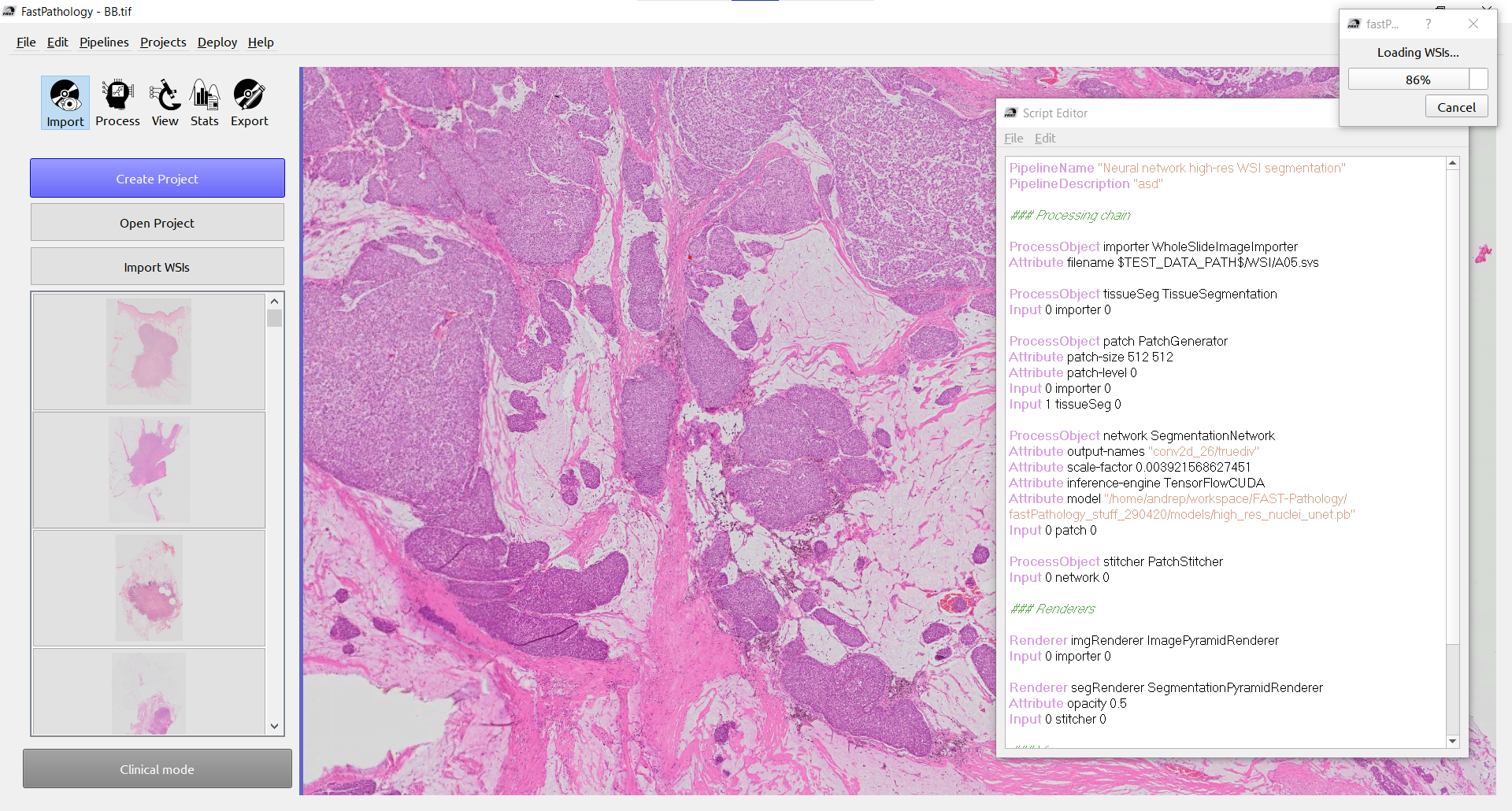}
    \caption{An example of FastPathology's GUI showing some basic functionalities. The task bar can be seen on the left side. The right side contains a OpenGL window rendering a WSI. On top of the window is a progress bar and a script editor containing a text pipeline.}
    \label{fig:example_GUI}
\end{figure*}

\subsection{Neural network processing} \label{neural network processing}
Inference of neural networks is done through FAST by loading a trained model stored on disk as described in \cite{8844665}.
FAST comes with multiple inference engines including: 1) Intel's OpenVINO which can run on Intel CPUs as well as their integrated GPUs, 2) Google's TensorFlow which can run on NVIDIA GPUs with the CUDA and cuDNN framework, or directly on CPUs, and 3) NVIDIA's TensorRT which can run on NVIDIA GPUs using CUDA and cuDNN.
FAST will automatically determine which inference engines can run on the current system depending on whether CUDA, cuDNN or TensorRT are installed or not.

In many image analysis solutions the WSI is tiled into small patches of a given size and magnification level. 
A method is then applied to each patch independently, and the results are stitched together to form the WSI's analysis result.
FAST uses a \textit{patch generator} to tile a WSI into patches in a separate thread on the CPU. Thus, a neural network can simultaneously process patches while new patches are being generated.
Due to the parallel nature of GPUs, it can be beneficial to perform neural network inference on batches of patches, which can be done in FAST using the \textit{patch to batch generator}.
Finally, the \textit{patch stitcher} in FAST takes the stream of patch-wise predictions to form a final result image or tensor which can be visualized or further analyzed.
For methods which generate objects such as bounding boxes, an \textit{accumulator} is used instead which simply concatenates the objects into a list.
Since computations and visualizations are run in separate threads in FAST, the predictions can be visualized on top of the WSI, while the patches are being processed. 

It is also possible to run different analyses on different threads in FAST.
However, as amount of memory and threads are limited, running multiple processes simultaneously might affect the overall runtime performance.

Results are stored differently depending on whether one is performing patch-wise classification, object detection or segmentation.
For patch-wise classification, predictions are visualized as small rectangles with different colors for different classes and varying opacity dependent on classification confidence level.
For object detection, predictions are visualized as bounding boxes, where the color of the box indicates the predicted class.
For semantic segmentation, pixels are classified, and given a color and opacity depending on the predicted class and confidence level.

To enable introduction of new models and generalizing to different multi-input/output network architectures, each model assumes that it has a corresponding \textit{model description text-file}. This file includes information on how the models should be handled. For instance, for some inference engines, the input size must be set by the user, as it is not interpretable directly from the model.

\subsection{Graphical user interface}
In order to use the WSI functionality in FAST without programming, a GUI is required. 
The GUI of FastPathology was implemented using Qt 5 \cite{Qt5}. 
The GUI was split into two windows. On the right side there is a large OpenGL window for visualizing WSIs and analysis results from CNN predictions.
On the left, the user can find a dynamic taskbar with five sections for handling WSIs.
\begin{enumerate}
    \item \textbf{Import:} Options to create or load existing projects and reading WSIs.
    \item \textbf{Process:} Selection of available processing methods, e.g. tissue segmentation or inference with CNN.
    \item \textbf{View:} Viewer for selecting results to visualize, e.g. tumor segmentation, patch-wise histological prediction.
    \item \textbf{Stats:} Extract statistics from results, e.g. histogram of histological grade predictions, final overall WSI-level prediction.
    \item \textbf{Export:} Exporting results in appropriate formats, e.g. .png or .mhd/.raw for segmentations and heatmaps, or .csv for inference results.
\end{enumerate}

\begin{figure*}[t!]
    \centering
    \includegraphics[width=1\textwidth]{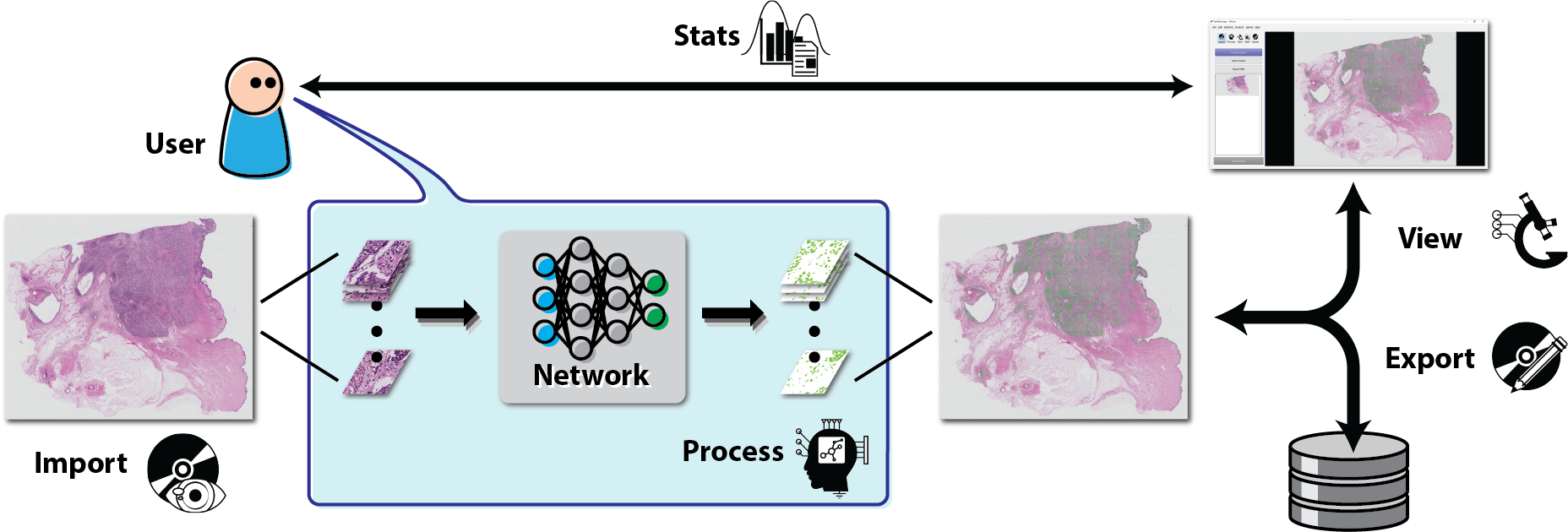}
    \caption{Illustration of the user workflow for analyzing WSIs in FastPathology. It also shows how each component in the GUI are related, and how the user can run a pipeline (Process) and get feedback from the neural network, either from the OpenGL window (View) or from the statistics summary (Stats). WSIs can be added through the Import widget and results are stored on disk using the Export widget.}
    \label{fig:pipeline}
\end{figure*}

An example of the GUI can be seen in \cref{fig:example_GUI}. The View, Stats and Export widgets are dynamically updated depending on which results are available.
In the View widget one can also change the opacity of the result or the class directly, and the color. Results can be removed and inference can be halted.
\cref{fig:pipeline} shows how the user can interact with the GUI and how the different components relate. 

\subsection{Text pipelines}
FAST implements \textit{text pipelines}, a txt-file containing information regarding which components to use in a pipeline.
These pipelines are deployable directly within the software.
It is also possible to load external pipelines, or to create or edit pipelines using the built-in script-editor, as seen in \cref{fig:example_GUI}.
To make the editor more user-friendly, text highlighting was added. This produces different colors for FAST objects and corresponding attributes, e.g. patch generator and magnification level.
Using FastPathology, it is also possible to modify other text-files, such as the model description text-file. 

\subsection{Advanced mode}
An advanced mode was added to enable users to change and tune hyperparameters of algorithms and models. For tissue segmentation, the threshold and kernel size for the morphological operators can be set in the GUI. A dynamic preview of the segmentation is then updated in real time, to give the user feedback about the selected parameters.

\subsection{Projects}
It may be convenient to run the same analysis on multiple WSIs. Therefore, a \textit{Project} can be created, and several WSIs can be added to the project. By selecting a pipeline, and choosing \textit{run-for-project}, the pipeline is run sequentially on all WSIs in the project. Results are stored within the project in a separate folder. This makes it possible to load the project including the results, and export the results to other platforms, e.g. QuPath.

\subsection{Storing results}
Storing results from different image analysis is an important part of a WSI analysis platform.
Currently, it is possible to store the tissue segmentation and predictions on disk using the metaimage (.mhd/.raw) format in FAST. Tensors from neural networks are stored using the HDF5 format.

\subsection{Inference use cases} \label{Inference use cases}
Four different neural network inference cases were selected to demonstrate the capabilities and performance of the application.
All models were implemented and trained using TensorFlow 1.13.
For use cases 1, 3 and 4, the tissue segmentation method was used to limit the neural network processing to tiles with tissue only.
All models were trained as a proof-of-concept for the platform, not to achieve the highest possible accuracy. 

\subsubsection{Use case 1 - Patch-wise classification}
\begin{figure*}[t!]
    \centering
    \includegraphics[width=1\textwidth]{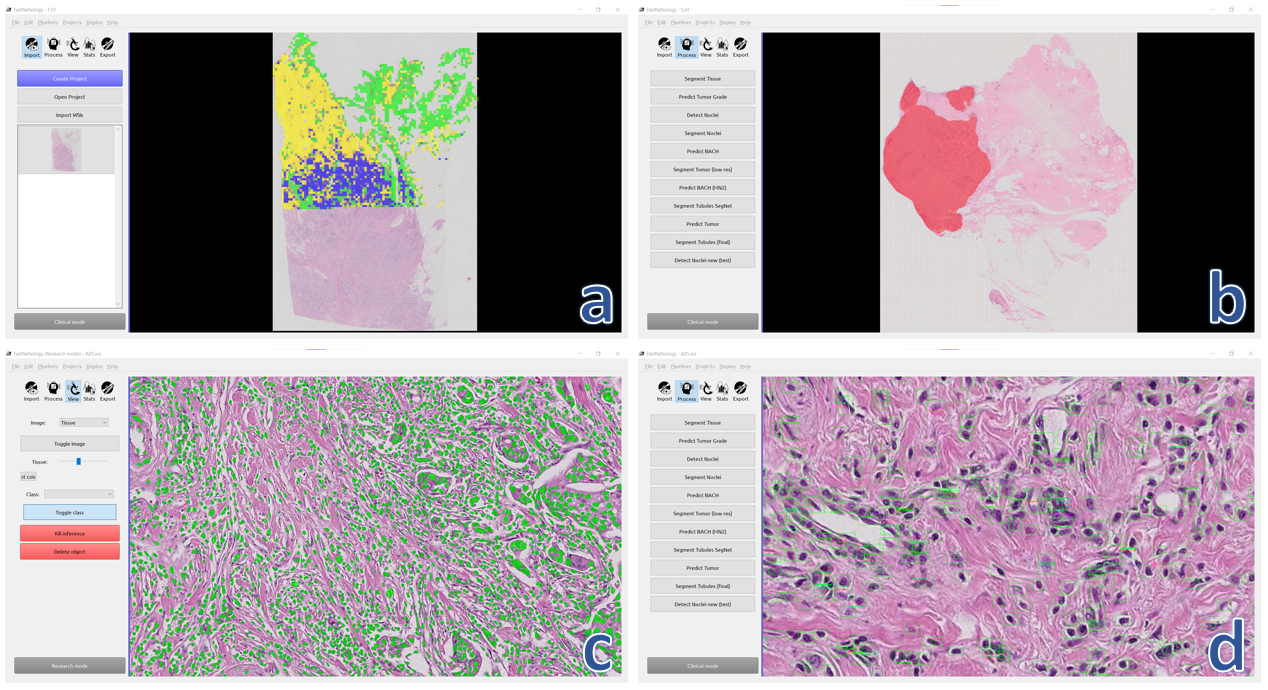}
    \caption{Illustrating the resulting predictions of each use case on top of a WSI. a) patch-wise classification of tissue, b) low-resolution segmentation of breast cancer tumor, c) high-resolution segmentation of cell nuclei, and d) object detection of cell nuclei.}
    \label{fig:subplot_use_cases}
\end{figure*}

This use case focuses on patch-wise classification of WSIs.
The image was tiled into non-overlapping tiles of size $512\times 512\times 3$ at $\times200$ magnification level, and RGB intensities normalized to the range $[0, 1]$.
The network used was a CNN with the MobileNetV2 \cite{MobileNetv2} encoder pre-trained on the ImageNet dataset \cite{deng2009imagenet}.
The classifier part contained a global average max pooling layer and two dense layers with 64 and 4 neurons respectively.
Between the dense layers, batch normalization, ReLU and dropout with a rate of 0.5 were used. In the last layer a softmax activation function was used to obtain the output probability prediction for each class.
The model has $\sim 2.31$M parameters.
It was trained on the Grand Challenge on Breast Cancer Histology Images (BACH) dataset \cite{Aresta2019}.
The model classifies tissue into four classes: normal tissue, benign lesion, in situ, and invasive carcinoma.
A patch stitcher is used to create a single heatmap of all the classified patches.
The heatmap is visualized on top of the WSI with a different color for each class. The opacity reflects the confidence score of the class as shown in \cref{fig:subplot_use_cases}a).

\subsubsection{Use case 2 - Low-resolution segmentation}
This task focuses on semantic segmentation of WSIs by segmenting pixels of the entire WSI using the pyramid level with the lowest resolution.
Thus, this use case does not process patches, but the entire image.
The network uses a fully convolutional encoder-decoder scheme, based on the U-Net architecture \cite{unet}.
From images of size $1024\times 1024\times 3$, the network classifies each pixel as tumor or background.
All the convolutional layers in the model are followed by batch normalization, ReLU and a spatial dropout of 0.1. However, in the the output layer, the softmax activation function was used.
The total number of parameters was $\sim 11.56$M.
The dataset used is a subset of a series of breast cancer cases curated by the Breast Cancer Subtypes Project \cite{Engstrom2013}. The subset comprises Hematoxylin-Eosin (H\&E)-stained full-face tissue sections ($4\mu$ thick) from breast cancer tumors. WSIs were captured at $\times400$ magnification.
The result is visualized on top of the WSI with each class having a different color. The opacity reflects the confidence score of the class. \cref{fig:subplot_use_cases}b) shows the results of this use case, where the segmented tumor region is shown in transparent red, whereas the background class is completely transparent.

\subsubsection{Use case 3 - High-resolution segmentation}
We used the same U-Net-architecture as in use case 2, to perform segmentation on independent patches.
The image was tiled as in use case 1. Tiles of size $256\times 256\times 3$ were used. Patches from varying image planes were extracted (around $\times200$), but higher resolution tiles were preferred.
The PanNuke dataset \cite{gamper2019pannuke,gamper2020pannuke} was used to train the model. PanNuke is a multi-organ pan-cancer dataset for nuclear segmentation and classification.
It contains 19 different tissue types and five different classes of nuclei: inflammatory cell, connective tissue, neoplastic, epithelial, and dead (apoptic or necrotic) nuclei.
We only trained the model to perform nuclear segmentation, regardless of class. 
The total number of parameters was $\sim 7.87$M.
The segmentation of each patch was stitched together to form a single, large segmentation image. This image has the same size as the image pyramid level it is processing, and the result is formed into a new image segmentation pyramid as described in section \ref{sec:creating_large_images}.
The result is visualized on top of the WSI with each class having a different color (see \cref{fig:subplot_use_cases}c).

\subsubsection{Use case 4 - Object detection and classification}
We used the same tiling strategy as for use case 3, with the same image planes and input size.
However, in this case we performed object detection using the Tiny-YOLOv3-architecture \cite{yolo3}.
Implementation and training of Tiny-YOLOv3 was inspired by the specified GitHub repository\footnote{\url{https://github.com/qqwweee/keras-yolo3}}.
The model was pretrained on the COCO dataset \cite{COCO2014}, and fine-tuned on the PanNuke dataset.
Bounding box coordinates with corresponding confidence and predicted class for all predicted candidates were made.
The total number of parameters was $\sim 8.67$M.
Non-maximum suppression was performed to handle overlapping bounding boxes. From all patches, these were then accumulated into one large bounding box set, visualized as colored lines with OpenGL, where the color indicates the predicted class (see \cref{fig:subplot_use_cases}d).

\section{Experiments} \label{Experiments}

\subsection{Runtime} \label{Runtime results}
To assess speed, we performed runtime experiments using the four use cases.
The experiments were run on a single Dell desktop with Ubuntu 18.04 64 bit operating system, with 32 GB of RAM, an Intel i7-9800X CPU and two NVIDIA GPUs, GeForce RTX 2070 and Quadro P5000.
We measured runtimes using the four inference engines: TensorFlow CPU, TensorFlow GPU (v1.14), OpenVINO CPU (v2020.3) and TensorRT (v7.0.0.11).
TensorRT was only used in use cases 1 and 4, where an UFF-model was available. All U-Net models contained spatial dropout and upsampling layers that were not supported by TensorRT, and thus could not be converted.
For each inference engine, a warmup run was done before 10 consecutive runs were performed. Runtimes for each module in a pipeline were reported. The warmup was done to avoid measurements being influenced by previous runs. The experiments were run sequentially.

From these experiments, the population mean ($\bar{X}$) and standard error of the mean ($S_{\bar{X}}$) were calculated.
Multiple Shapiro-Wilk tests \cite{ShapiroWilk1965} were conducted to state whether the data were normal. The Benjamini-Hochberg false discovery rate method \cite{FDR1995} was used to correct for multiple testing. For all hypothesis tests, a significance level of 5 \% was used. Only six out of 32 variables had small deviations from the normal distribution, thus a normal distribution was assumed.
The mean and 95\%-confidence intervals were reported. 
In addition, multiple pairwise tests were performed using Tukey's range test \cite{Tukey1949ComparingIM} to evaluate whether there were a significant difference between any of the total runtimes (see supplementary material for the p-values).

All experiments were done on the A05.svs $\times200$ WSI from the BACH dataset. Measurements were in milliseconds, if not stated otherwise.
To simplify the measurements, rendering was excluded in all runtime measurements. The OpenGL rendering runtime is so small it can be regarded as negligible. The real bottleneck is inference speed and patch generation.

For all runtime measurements we reported the time used for each component (patch generator, neural network input and output processing, neural network inference, and patch stitcher), and the combined time in a FAST pipeline. 
Neural network input processing includes resizing the images if necessary and intensity normalization (0-255 $\rightarrow$ 0-1).

\subsection{Memory}
We monitored memory usage on selected tasks and compared them to the QuPath (v0.2.3), ASAP (v1.9) and Orbit (v3.64) platforms.
All experiments were run on the same Dell desktop as used in \cref{Runtime results} (using the RTX 2070 GPU).
The WSI used was the TE-014.svs from the Tumor Proliferation Assessment Challenge 2016 \cite{TUPAC16}, since it is a large, openly available $\times400$ WSI.

In this experiment, memory usage was measured after starting the program, after opening the WSI, and after zooming and panning the view for 2.5 minutes. 
Both RAM and GPU memory usage was measured. 
To make the comparison fair, we attempted to make similar movements and zoom for all platforms.

Orbit initializes the WSI from a zoomed region, in contrast to the three other which initializes from a low-resolution overview image. In order to achieve the same overview field of view for all, it was necessary to zoom out initially when using Orbit. This, however spiked the RAM usage for Orbit. Thus, to make comparison fair, we only measured memory usage after the initial image was displayed when opening a WSI for all applications.

The physical memory usage was monitored using the interactive process viewer \textit{htop} on Linux. Due to this, if a process used more than 10 GB of RAM, htop would report it as 0.001 TB, which meant that we had lower resolution on these measurements. The graphical memory usage was monitored using the NVIDIA System Mamagement Interface (\textit{nvidia-smi}).

\subsection{Model and hardware choice}
To further assess how different neural network architectures could affect inference speed on a specific use case, we ran use case 1 with a more demanding InceptionV3 model \cite{8844665}. This model is available in the FAST test data\footnote{\url{https://github.com/smistad/FAST/wiki/Test-data}}.
The model should have a classifier part identical to the one used for our MobileNetV2 model.

The same use case was also run on a low-end HP laptop with Windows 10 64 bit operating system, 16 GB of RAM, an Intel i7-7600 CPU, and Intel HD Graphics 620 integrated GPU, to show how runtimes could differ between low- and high-end machines. 

To compare difference in runtime between operating systems, we also run the same experiments using a high-end Razer laptop with Windows 10 64 bit operating system, 32 GB of RAM, an Intel i7-10750H CPU, an Intel UHD graphics integrated GPU, and NVIDIA GeForce RTX 2070 Max-Q GPU. To our understanding, the performance of both the CPU and GPU should be comparable to that of the Dell desktop computer used in the experiments. During experiments with both Windows laptops, the machines were constantly being charged and real-time anti-malware protection was turned off.
For all machines, all experiments were performed using a Solid State Drive (SSD).


\section{Results} \label{Results}

\subsection{Runtime}
Comparing the choice of inference engine, \cref{tab:inference_speed_case1,tab:inference_speed_case2,tab:inference_speed_case3,tab:inference_speed_case4} show that inference with TensorFlow CPU was the slowest alternative, for each respective use case, especially using TensorFlow CPU (see supplementary material for the p-values).
Inference with GPU was the fastest, with TensorRT slightly faster than TensorFlow CUDA. However, no significant difference was found between TensorFlow CUDA and TensorRT in any of the runtime experiments.
The OpenVINO CPU IE had comparable inference speed with the GPU alternatives, even surpassing TensorFlow CUDA on the low-resolution segmentation task. However, no significant difference was observed.
Thus, there was no benefit of using the GPU for low-resolution segmentation. 
We also ran inference with two different GPUs using TensorRT, and found negligible difference in terms of inference speed between the two hardwares.
Also, more complex tasks such as object detection and high-resolution segmentation resulted in slower runtimes than patch-wise classification and low-resolution segmentation, across all inference engines.


\begin{table*}[b!]
    \centering
    \renewcommand{\arraystretch}{1.5}
    \caption{Memory measurements of reading, panning and zooming the view of a $\times400$ WSI. All memory usage values are in MB.}
    \begin{tabular}{lccccccccc}
        \toprule
        \textbf{Memory usage} & \multicolumn{2}{c}{\textbf{FastPathology}} & \multicolumn{2}{c}{\textbf{QuPath}} & \multicolumn{2}{c}{\textbf{Orbit}} & \multicolumn{2}{c}{\textbf{ASAP}} \\\cmidrule(lr){2-3}\cmidrule(lr){4-5}\cmidrule(lr){6-7}\cmidrule(lr){8-9} & RAM & GPU & RAM & GPU & RAM & GPU & RAM & GPU \\ \midrule
        \textbf{Application startup} & 205 & 101 & 497 & 86 & 373 & 0 & 84 & 0 \\
        \textbf{Opening WSI} & 268 & 111 & 989 & 88 & 817 & 0 & 173 & 0 \\
        \textbf{Zooming and panning} &  1,544 & 1,203 & $\sim$11,000 & 89 & 9,903 & 0 & 1,185 & 0 \\
        \bottomrule
    \end{tabular}
    \label{tab:memory_measurements40x}
\end{table*}

\begin{table*}[p]
    \centering
    \renewcommand{\arraystretch}{1.5}
    \caption{Runtime measurements of use case 1 - Patch-wise classification, using the MobileNetV2 encoder performed on the Ubuntu desktop. Each row corresponds to an experimental setup. Each cell displays the average runtime and 95 \% confidence interval limits for 10 successive runs.}
    \adjustbox{width=\textwidth}{
    \begin{tabular}{llcccccc}
        \toprule
        \multirow{2}{*}{\textbf{Inference engine}} & \multirow{2}{*}{\textbf{Processor}} & \multicolumn{6}{c}{\textbf{Runtime (ms)}} \\ \cmidrule(rl){3-8}
        & & Patch generator & NN input & NN inference & NN output & NN patch stitcher & Total (s) \\
        \midrule
        OpenVINO CPU & Intel i7-9800X & $29.7 \pm 0.0$ & $1.4 \pm 0.0$ & $16.7 \pm 0.0$ & $0.0 \pm 0.0$ & $ 0.0 \pm 0.0$ & $145.0 \pm 0.2$\\
        TensorFlow CPU & Intel i7-9800X & $34.0 \pm 0.0$ & $1.1 \pm 0.0$ & $35.6 \pm 0.4$ & $0.0 \pm 0.0$ & $0.0 \pm 0.0$ & $176.4 \pm 2.0$  \\
        TensorFlow CUDA & Quadro P5000 & $21.3 \pm 0.4$ & $1.5 \pm 0.0$ & $9.3 \pm 0.3$ & $0.0 \pm 0.0$ & $0.0 \pm 0.0$ & $103.8 \pm 2.2$ \\
        TensorRT & GeForce RTX 2070 & $20.2 \pm 0.1$ & $1.3 \pm 0.0$ & $1.2 \pm 0.0$ & $0.0 \pm 0.0$ & $0.0 \pm 0.0$ & $98.9 \pm 0.5$ \\
        & Quadro P5000 & $20.4 \pm 0.1$ & $1.3 \pm 0.0$ & $1.3 \pm 0.1$ & $0.0 \pm 0.0$ & $0.0 \pm 0.0$ & $99.8 \pm 0.6$ \\
        \bottomrule
    \end{tabular}
    }
    \label{tab:inference_speed_case1}
\centering
\renewcommand{\arraystretch}{1.5}
\caption{Runtime measurements of use case 2 - Low-resolution semantic segmentation}
\adjustbox{width=\textwidth}{
\begin{tabular}{llccccc}
    \toprule
    \multirow{2}{*}{\textbf{Inference engine}} & \multirow{2}{*}{\textbf{Processor}} & \multicolumn{5}{c}{\textbf{Runtime (ms)}} \\ \cmidrule(rl){3-7}
    & & Image read & NN input & NN inference & NN output & Total (s) \\
    \midrule
    OpenVINO CPU & Intel i7-9800X & $9.4 \pm 3.7$ & $5.3 \pm 0.1$ & $149.3 \pm 3.1$ & $4.6 \pm 5.7$ & $0.17 \pm 0.01$ \\
    TensorFlow CPU & Intel i7-9800X & $8.5 \pm 0.7$ & $3.1 \pm 0.1$ & $1101.9 \pm 11.5$ & $2.7 \pm 0.1$ & $1.12 \pm 0.01$  \\
    TensorFlow CUDA & Quadro P5000 & $10.7 \pm 1.7$ & $3.3 \pm 0.2$ & $998.9 \pm 12.8$ & $7.2 \pm 1.3$ & $1.0 \pm 0.0$ \\
    \bottomrule
\end{tabular}
}
\label{tab:inference_speed_case2}
    \centering
    \renewcommand{\arraystretch}{1.5}
    \caption{Runtime measurements of use case 3 - High-resolution semantic segmentation}
    \adjustbox{width=\textwidth}{
    \begin{tabular}{llcccccc}
        \toprule
        \multirow{2}{*}{\textbf{Inference engine}} & \multirow{2}{*}{\textbf{Processor}} & \multicolumn{6}{c}{\textbf{Runtime (ms)}} \\ \cmidrule(rl){3-8}
        & & Patch generator & NN input & NN inference & NN output & NN patch stitcher & Total (s) \\
        \midrule
        OpenVINO CPU & Intel i7-9800X & $37.6 \pm 0.1$ & $0.8 \pm 0.0$ & $16.2 \pm 0.0$ & $0.2 \pm 0.0$ & $66.3 \pm 0.1$ & $400.7 \pm 0.7$ \\
        TensorFlow CPU & Intel i7-9800X & $37.3 \pm 0.2$ & $1.0 \pm 0.0$ & $38.5 \pm 0.2$ & $0.3 \pm 0.0$ & $73.2 \pm 0.1$ & $542.3 \pm 1.2$ \\
        TensorFlow CUDA & Quadro P5000 & $29.9 \pm 0.5$ & $0.8 \pm 0.0$ & $5.3 \pm 0.0$ & $0.3 \pm 0.0$ & $76.1 \pm 0.3$ & $396.2 \pm 1.6$ \\
        \bottomrule
    \end{tabular}
    }
    \label{tab:inference_speed_case3}
    \centering
    \renewcommand{\arraystretch}{1.5}
    \caption{Runtime measurements of use case 4 - Object detection and classification}
    \adjustbox{width=\textwidth}{
    \begin{tabular}{llcccccc}
        \toprule
        \multirow{2}{*}{\textbf{Inference engine}} & \multirow{2}{*}{\textbf{Processor}} & \multicolumn{6}{c}{\textbf{Runtime (ms)}} \\ \cmidrule(rl){3-8}
        & & Patch generator & NN input & NN inference & NN output & NN patch stitcher & Total (s) \\
        \midrule
        OpenVINO CPU & Intel i7-9800X & $9.9 \pm 0.0$ & $0.9 \pm 0.0$ & $6.7 \pm 0.1$ & $0.1 \pm 0.0$ & $0.0 \pm 0.0$ & $193.5 \pm 0.5$ \\
        TensorFlow CPU & Intel i7-9800X & $10.6 \pm 0.2$ & $1.2 \pm 0.1$ & $14.3 \pm 0.2$ & $0.0 \pm 0.0$ & $0.0 \pm 0.0$ & $295.0 \pm 3.0$ \\
        TensorFlow CUDA & Quadro P5000 & $6.6 \pm 0.0$ & $1.1 \pm 0.0$ & $3.4 \pm 0.0$ & $0.0 \pm 0.0$ & $0.0 \pm 0.0$ & $129.7 \pm 0.4$ \\
        TensorRT & Quadro P5000 & $6.6 \pm 0.0$ & $1.1 \pm 0.0$ & $1.0 \pm 0.0$ & $0.0 \pm 0.0$ & $0.0 \pm 0.0 $ & $129.4 \pm 0.3$ \\
        \bottomrule
    \end{tabular}
    }
    \label{tab:inference_speed_case4}
\end{table*}

\begin{table*}[p]
    \centering
    \renewcommand{\arraystretch}{1.5}
    \caption{Runtime measurements of patch-wise classification (use case 1), using the InceptionV3 encoder performed on the Ubuntu desktop.}
    \adjustbox{width=\textwidth}{
    \begin{tabular}{llcccccc}
        \toprule
        \multirow{2}{*}{\textbf{Inference engine}} & \multirow{2}{*}{\textbf{Processor}} & \multicolumn{6}{c}{\textbf{Runtime (ms)}} \\ \cmidrule(rl){3-8}
        & & Patch generator & NN input & NN inference & NN output & NN patch stitcher & Total (s) \\
        \midrule
        OpenVINO CPU & Intel i7-9800X & $28.4 \pm 0.1$ & $1.2 \pm 0.0$ & $49.9 \pm 0.1$ & $0.0 \pm 0.0$ & $0.0 \pm 0.0$ & $245.4 \pm 0.3$ \\
        TensorFlow CPU & Intel i7-9800X & $34.9 \pm 0.0$ & $1.3 \pm 0.0$ & $53.5 \pm 0.0$ & $0.0 \pm 0.0$ & $0.0 \pm 0.0$ & $263.0 \pm 0.2$ \\
        TensorFlow CUDA & Quadro P5000 & $21.2 \pm 0.1$ & $1.3 \pm 0.0$ & $23.3 \pm 0.1$ & $0.0 \pm 0.0$ & $0.0 \pm 0.0$ & $118.8 \pm 0.4$ \\
        \bottomrule
    \end{tabular}
    }
    \label{tab:inference_speed_case1_inceptionv3}
    \centering
    \renewcommand{\arraystretch}{1.5}
    \caption{Runtime measurements of patch-wise classification (use case 1), using the MobileNetV2 encoder performed on the low-end Windows laptop.}
    \adjustbox{width=\textwidth}{
    \begin{tabular}{llcccccc}
        \toprule
        \multirow{2}{*}{\textbf{Inference engine}} & \multirow{2}{*}{\textbf{Processor}} & \multicolumn{6}{c}{\textbf{Runtime (ms)}} \\ \cmidrule(rl){3-8}
        & & Patch generator & NN input & NN inference & NN output & NN patch stitcher & Total (s) \\
        \midrule
        OpenVINO CPU & Intel i7-7600U & $51.3 \pm 2.3$ & $3.6 \pm 0.1$ & $51.8 \pm 2.4$ & $0.0 \pm 0.0$ & $0.0 \pm 0.0$ & $268.4 \pm 12.2$ \\
        OpenVINO GPU & Intel HD Graphics 620 & $63.9 \pm 0.9$ & $4.3 \pm 0.0$ & $28.6 \pm 0.2$ & $0.0 \pm 0.0$ & $0.0 \pm 0.0$ & $314.0 \pm 4.1$ \\
        TensorFlow CPU & Intel i7-7600U & $104.9 \pm 3.9$ & $4.1 \pm 0.1$ & $218.2 \pm 2.1$ & $0.0 \pm 0.0$ & $0.0 \pm 0.0$ & $1065.4 \pm 10.2$ \\
        \bottomrule
    \end{tabular}
    }
    \label{tab:runtime_case1_windows}
    \centering
    \renewcommand{\arraystretch}{1.5}
    \caption{Runtime measurements of patch-wise classification (use case 1), using the InceptionV3 encoder performed on the low-end Windows laptop.}
    \adjustbox{width=\textwidth}{
    \begin{tabular}{llcccccc}
        \toprule
        \multirow{2}{*}{\textbf{Inference engine}} & \multirow{2}{*}{\textbf{Processor}} & \multicolumn{6}{c}{\textbf{Runtime (ms)}} \\ \cmidrule(rl){3-8}
        & & Patch generator & NN input & NN inference & NN output & NN patch stitcher & Total (s) \\
        \midrule
        OpenVINO CPU & Intel i7-7600U & $48.6 \pm 0.3$ & $3.6 \pm 0.0$ & $299.2 \pm 1.5$ & $0.0 \pm 0.0$ & $0.0 \pm 0.0$ & $1449.4 \pm 7.4$ \\
        OpenVINO GPU & Intel HD Graphics 620 & $159.6 \pm 0.4$ & $5.7 \pm 0.0$ & $153.2 \pm 0.2$ & $0.0 \pm 0.0$ & $0.0 \pm 0.0$ & $788.6 \pm 2.0$ \\
        TensorFlow CPU & Intel i7-7600U & $100.62 \pm 0.9$ & $4.6 \pm 0.0$ & $448.4 \pm 1.8$ & $0.0 \pm 0.0$ & $0.0 \pm 0.0$ & $2167.3 \pm 8.9$ \\
        \bottomrule
    \end{tabular}
    }
    \label{tab:runtime_case1_inceptionv3_windows}
    \centering
    \renewcommand{\arraystretch}{1.5}
    \caption{Runtime measurements of patch-wise classification (use case 1), using the MobileNetV2 encoder performed on the high-end Windows laptop.}
    \adjustbox{width=\textwidth}{
    \begin{tabular}{llcccccc}
        \toprule
        \multirow{2}{*}{\textbf{Inference engine}} & \multirow{2}{*}{\textbf{Processor}} & \multicolumn{6}{c}{\textbf{Runtime (ms)}} \\ \cmidrule(rl){3-8}
        & & Patch generator & NN input & NN inference & NN output & NN patch stitcher & Total (s) \\
        \midrule
        OpenVINO CPU & Intel i7-10750H & $31.6 \pm 0.4$ & $2.2 \pm 0.0$ & $22.5 \pm 0.1$ & $0.0 \pm 0.0$ & $0.0 \pm 0.0$ & $155.3 \pm 1.8$ \\
        OpenVINO GPU & Intel UHD graphics & $37.4 \pm 0.2$ & $2.5 \pm 0.0$ & $28.3 \pm 0.1$ & $0.0 \pm 0.0$ & $0.0 \pm 0.0$ & $184.0 \pm 0.8$ \\
        TensorFlow CPU & Intel i7-10750H & $48.0 \pm 0.1$ & $2.4 \pm 0.0$ & $79.9 \pm 0.2$ & $0.0 \pm 0.0$ & $0.0 \pm 0.0$ & $395.1 \pm 1.2$ \\
        TensorRT & RTX 2070 Max-Q & $21.8 \pm 0.1$ & $2.2 \pm 0.0$ & $5.0 \pm 0.0$ & $0.0 \pm 0.0$ & $0.0 \pm 0.0$ & $108.5 \pm 0.4$ \\
        \bottomrule
    \end{tabular}
    }
    \label{tab:runtime_case1_high_end_windows}
    \centering
    \renewcommand{\arraystretch}{1.5}
    \caption{Runtime measurements of patch-wise classification (use case 1), using the InceptionV3 encoder performed on the high-end Windows laptop.}
    \adjustbox{width=\textwidth}{
    \begin{tabular}{llcccccc}
        \toprule
        \multirow{2}{*}{\textbf{Inference engine}} & \multirow{2}{*}{\textbf{Processor}} & \multicolumn{6}{c}{\textbf{Runtime (ms)}} \\ \cmidrule(rl){3-8}
        & & Patch generator & NN input & NN inference & NN output & NN patch stitcher & Total (s) \\
        \midrule
        OpenVINO CPU & Intel i7-10750H & $32.8 \pm 0.7$ & $2.4 \pm 0.0$ & $118.2 \pm 0.3$ & $0.0 \pm 0.0$ & $0.0 \pm 0.0$ & $578.6 \pm 1.4$ \\ 
        OpenVINO GPU & Intel UHD graphics & $33.7 \pm 0.1$ & $2.8 \pm 0.0$ & $111.3 \pm 0.1$ & $0.0 \pm 0.0$ & $0.0 \pm 0.0$ & $547.3 \pm 0.3$ \\
        TensorFlow CPU & Intel i7-10750H & $47.2 \pm 0.2$ & $2.5 \pm 0.0$ & $165.7 \pm 0.6$ & $0.0 \pm 0.0$ & $0.0 \pm 0.0$ & $805.6 \pm 3.0$ \\
        TensorRT & RTX 2070 Max-Q & $22.2 \pm 0.1$ & $2.1 \pm 0.0$ & $11.2 \pm 0.0$ & $0.0 \pm 0.0$ & $0.0 \pm 0.0$ & $110.6 \pm 0.5$ \\
        \bottomrule
    \end{tabular}
    }
    \label{tab:runtime_case1_inceptionv3_high_end_windows}
\end{table*}

\subsection{Memory}
With regards to memory, there was a strong difference between the C\textit{++} and the Java-based applications (see \cref{tab:memory_measurements40x}).
Both C\textit{++}-based platforms used considerably less memory across all experiments. 
Using nvidia-smi we observed that FastPathology was the only platform that ran both computation and graphics on the GPU (C+G). FAST uses OpenCL for computations and OpenGL for rendering. The two Java-based softwares (QuPath and Orbit) only ran graphics on GPU, either using DirectX or another non-OpenGL form of rendering.
ASAP and Orbit did not use any GPU, whereas QuPath used a negligible amount. Hence, FastPathology was the only platform capable of exploiting the advantage of having a GPU available for both computations and rendering.
It was observed that both C\textit{++} applications (FastPathology and ASAP) opened their WSIs almost instantly, whereas both Java-based softwares (QuPath and Orbit) took a few seconds.

\subsection{Model and hardware choice}
\cref{tab:inference_speed_case1,tab:inference_speed_case1_inceptionv3} show runtime measurements on use case 1 using two different networks, MobileNetV2 and InceptionV3. Both networks are commonly used in digital pathology \cite{Aresta2019,Kassani2019,Skrede2020}, but the latter is more computationally demanding. Thus, inference with InceptionV3 was slower overall than with MobileNetV2. However, due to the increase in complexity, we observed that inference using CUDA was faster than using all CPU alternatives, in use case 1. This example showed that having a GPU available for inference can greatly speed up runtime, especially when models become more complex. 
A similar conclusion can also be drawn from \cref{tab:inference_speed_case4} where a complex U-Net architecture was used, in contrast to using a lightweight Tiny-YOLOv3 architecture as seen in \cref{tab:inference_speed_case3}. 

\cref{tab:runtime_case1_windows,tab:runtime_case1_inceptionv3_windows} show inference using the low-end laptop. There was a significant increase in runtime for all inference engines. The low-end laptop had an integrated GPU and thus we could run inference using OpenVINO GPU. This alternative is only better when more demanding models are used. Here, a much larger difference in runtime can be seen between the two CPU alternatives, TensorFlow CPU and OpenVINO CPU. OpenVINO was superior in terms of runtime.

\cref{tab:runtime_case1_high_end_windows,tab:runtime_case1_inceptionv3_high_end_windows} show runtime measurements of the same use case with both encoders using the high-end Windows laptop. In this case we achieved runtime performance similar to the performance using the Ubuntu desktop. We found no significant difference using TensorRT between the two high-end machines, and TensorRT on Windows and TensorFlow CUDA on Ubuntu. For CPU there was a significant drop in performance for all use cases and encoders.

\section{Discussion}

In this paper, we have presented a new platform, FastPathology, for visualization and analysis of WSIs. We have described the components developed to achieve this high-performance and easy-to-use platform. The software was evaluated in terms of memory usage, inference speed, and model and OS compatibility (see supplementary material for the p-values). A variety of deep learning use cases, model architectures, inference engines and processors were used. 

\subsection{Memory usage and runtime}
In the memory experiments, FastPathology performed similarly to another C\textit{++}-based software (ASAP), whereas both Java-based alternatives (QuPath and Orbit) were more memory intensive, using a large amount of memory while zooming. We have presented a runtime benchmark. Among the CPU alternatives, OpenVINO CPU performed the best. Inference on GPU was the fastest, but no significant difference was found when comparing TensorFlow CUDA and TensorRT. 
A small degradation in runtime was observed when using Windows compared to Ubuntu, but there was no significant difference using GPU. Runtimes on the low-end machine were slower, especially for more demanding models, but if an integrated GPU is available, inference can be improved using OpenVINO GPU.

In use case 2, OpenVINO outperformed TensorFlow, even with GPU. This may be due to TensorFlow having a larger overhead compared to OpenVINO, which can clearly be seen when comparing against the TensorFlow CPU alternative. 
For TensorFlow CUDA, CUDA initialization was included in the inference runtime, which is why OpenVINO CPU \textit{appear} to be faster than the GPU alternative. However, this initialization penalty will only affect the first patch, as CUDA is cached for all new patches.
Whereas in use case 1, using TensorRT, we achieved similar runtime using two GPUs with quite varying memory size, 16 GB vs. 8 GB. This can be explained by both GPUs having similar computational power, and FAST only using the memory required to perform the task at hand. Hence, having a GPU with more memory does not necessarily improve runtime.


A slower runtime on the low-end machine can be explained by a lower frequency and number of cores (2 vs. 6) of the CPU. FAST takes advantage of all cores during inference and visualization. Thus, having a greater number of cores is beneficial, especially when running inference in parallel. Using the high-end machine on Windows, we also saw a small degradation in runtime using CPU. This may be explained by Windows having larger overhead compared to Ubuntu, or differences in hardware components that were not considered in this study, e.g. SSD.
However, on GPU using TensorRT, there was a negligible difference between the two high-end machines. The small drop in performance might be due to the Windows machine having a Max-Q GPU design which is known to slightly limit the performance of the GPU, especially with regards to speed.

Even though the CPU alternatives have a longer total runtime than the GPU alternatives, this cannot be explained by the higher inference speed on GPU alone. In FAST, the patch generation happens in parallel on the CPU. If a GPU is available, the CPU can focus on generating patches, while the GPU performs inference simultaneously. 
This optimizes the pipeline, as reading of patches is slow due to the virtual memory mapping of the WSI. If no GPU is available, the CPU must perform both tasks, and with limited cores, the overall runtime will increase.

\subsection{Comparison with other platforms}
QuPath is known to have a responsive, user-friendly viewer, with a seamless rendering of patches from different magnification levels. An optimized memory management or allocation of large amounts of data in memory is required to provide such a user experience. This could explain why QuPath used the largest amount of memory of all four tested solutions. FastPathology and ASAP provide a similar experience with a considerably smaller memory footprint. 
Rendering WSIs with Orbit did not work as swiftly, neither on Ubuntu nor Windows. 

There is a wide range of platforms to choose from when working with WSIs. Solutions such as ASAP are made to be lightweight and responsive in order to support visualization and annotation of giga-resolution WSIs. Platforms such as QuPath enable deployment of built-in image analysis methods, either in Groovy, Python, or through ImageJ, as well as the option to implement the user's own methods. Orbit takes it further by making it possible for the user to train and deploy their own deep learning models in Python within the software. 
FastPathology can deploy CNNs in the same way as Orbit while maintaining comparable memory consumption to ASAP during visualization. It is also simple and user-friendly, requiring no code-interaction to deploy models.


Some models are more demanding and thus naturally require greater memory. To some extent, memory usage can be adjusted through pipeline design, and by choice of model compression and inference engine.
Depending on the hardware, FastPathology takes advantage of all availble resources to produce a tailored experience when deploying models. Thus, pipeline designs such as batch inference can be done to further improve runtime performance.
However, it is also possible to deploy models on low-end machines, even without a GPU.
Machines that are using Intel CPUs, typically also include integrated graphics. In this case OpenVINO GPU could be used to improve runtime performance.

In FastPathology, the components used for reading, rendering and processing WSIs, and displaying predictions on top of the image, are made available through FAST. 
Since Python is one of the most popular languages for data scientists to develop neural network methods, FAST has been made available in Python as an official pip package\footnote{\textit{pip install pyfast} - \url{https://pypi.org/project/pyFAST/}}, and is currently available for Ubuntu (version 18 and 20) and Windows 10, with OpenVINO and TensorRT included as inference engines.
This means that platforms that can use Python (e.g. Orbit and QuPath), could also use our solutions, for instance for enabling or improving deployment of CNNs. 



\subsection{Strengths and weaknesses}

The platform has been developed through close collaboration with the pathologists at St. Olavs Hospital, Trondheim, Norway, to ensure user-friendliness and clinical relevance. The memory usage of the platform for reading, visualizing and panning a $\times400$ WSI has been compared to three existing softwares. As none of the existing platforms have published runtime benchmarks, we performed a thorough study to produce a benchmark. The WSIs, models and source code for running these experiments have been made public to facilitate reproducibility and encourage others to run similar benchmarks.

The runtime measurements were only performed using three machines. Runtimes on new machines may vary, depending on hardware, as well as version and configuration of the OS. It is possible to further improve runtime by compressing models (e.g. using half precision), using a different patch size, or running models on lower magnification levels. However, this might degrade the final result. Such a study would require a more in-depth analysis in the trade-off between design and performance. As the models used were only trained to show proof of concept, this was considered outside the scope of this paper.

Regarding memory usage, experiments were only run \textit{once} on \textit{one} machine as these experiments were performed manually and were tedious to repeat. The measurements were performed by one person, and not the most likely end-user of the platform. 
Thus, in the future, a more in-depth study should be done to verify to what extent runtime and memory consumption differ depending on OS, hardware and user-interaction with the viewer. Including memory usage for the use cases would be interesting. However, FastPathology is the only platform to stream CNN-based predictions as overlays during inference, and thus a fair comparison cannot be made. 

FastPathology is continuously in development, and thus this paper only presents the first release. 
Future work includes support for more complex models, support for more WSI and neural network storage formats, and basic annotation and region of interest tools.
As this is an open-source project, we encourage the community to contribute through GitHub.

\section{Conclusion} \label{Conclusion}

In this paper, we presented an open-source deep learning-based platform for digital pathology called FastPathology. It was implemented in C\texttt{++} using the FAST framework, and was evaluated in terms of runtime on four use cases, and in terms of memory usage while viewing a $\times400$ WSI. FastPathology had comparable memory usage compared to another C\textit{++} platform, outperforming two Java-based platforms. In addition, FastPathology was the only platform that can perform neural network predictions and visualize the results as overlays in real-time, as well as having a user-friendly way of deploying external models, access to a variety of different inference engines, and utilize both CPU and GPU for rendering and processing. Source code, binary releases and test data can be found online on GitHub at \url{https://github.com/SINTEFMedtek/FAST-Pathology/}.

\bibliographystyle{IEEEtran}
\bibliography{mendeley}

\end{document}